\documentclass[runningheads]{llncs}
\usepackage{multirow} 
\usepackage[T1]{fontenc}
\usepackage{graphicx,verbatim}
\begin{document}
\title{Unlabeled Cross-Center Automatic Analysis for TAAD: An Integrated Framework from Segmentation to Clinical Features }
%

\author{Mengdi Liu, Qiang Li, Weizhi Nie$^{\ast}$, Shaopeng Zhang, and Yuting Su}  
\institute{Tianjin University \\}

  
\maketitle              
\begin{abstract}
Type A Aortic Dissection (TAAD) is a life-threatening cardiovascular emergency that demands rapid and precise preoperative evaluation. While key anatomical and pathological features are decisive for surgical planning, current research focuses predominantly on improving segmentation accuracy, leaving the reliable, quantitative extraction of clinically actionable features largely under-explored. Furthermore, constructing comprehensive TAAD datasets requires labor-intensive, expert-level pixel-wise annotations, which is impractical for most clinical institutions. Due to significant domain shift, models trained on a single-center dataset also suffer from severe performance degradation during cross-institutional deployment. This study addresses a clinically critical challenge: the accurate extraction of key TAAD clinical features during cross-institutional deployment in the total absence of target-domain annotations. To this end, we propose an unsupervised domain adaptation (UDA)-driven framework for the automated extraction of TAAD clinical features. The framework leverages limited source-domain labels while effectively adapting to unlabeled data from target domains. Tailored for real-world emergency workflows, our framework aims to achieve stable cross-institutional multi-class segmentation, reliable and quantifiable clinical feature extraction, and practical deployability independent of high-cost annotations. Extensive experiments demonstrate that our method significantly improves cross-domain segmentation performance compared to existing state-of-the-art approaches. More importantly, a reader study involving multiple cardiovascular surgeons confirms that the automatically extracted clinical features provide meaningful assistance for preoperative assessment, highlighting the practical utility of the proposed end-to-end segmentation-to-feature pipeline.

\keywords{Type-A Aortic Dissection  \and Clinically-oriented Evaluation \and Medical Image Segmentation.}

\end{abstract}
\section{Introduction}
Type-A Aortic Dissection (TAAD) is a hyperacute cardiovascular emergency characterized by rapid progression and high mortality\cite{zhu2020type}. The mortality rate increases by 1\% per hour following onset, exceeding 50\% if surgical intervention is not performed within 72 hours\cite{criado2011aortic}. Within this narrow therapeutic window, surgeons must conduct rapid and precise preoperative assessments via Contrast-Enhanced Computed Tomography (CECT)\cite{kesieme2024recognition}. Key clinicopathological features—including the Location of Tear (LOT), True/False Lumen (TL/FL) relationship, Branch Vessel Involvement (BVI), True Lumen Collapse (TLC), and False Lumen Area Ratio (FLAR)—serve as the critical foundation for determining surgical incisions, aortic arch management strategies, stent sizing, and risk stratification\cite{furui2024relationship,wang2023prognostic,igarashi2022ratio}.  

\begin{figure*}[!t]
\centering
\includegraphics[width=\textwidth]{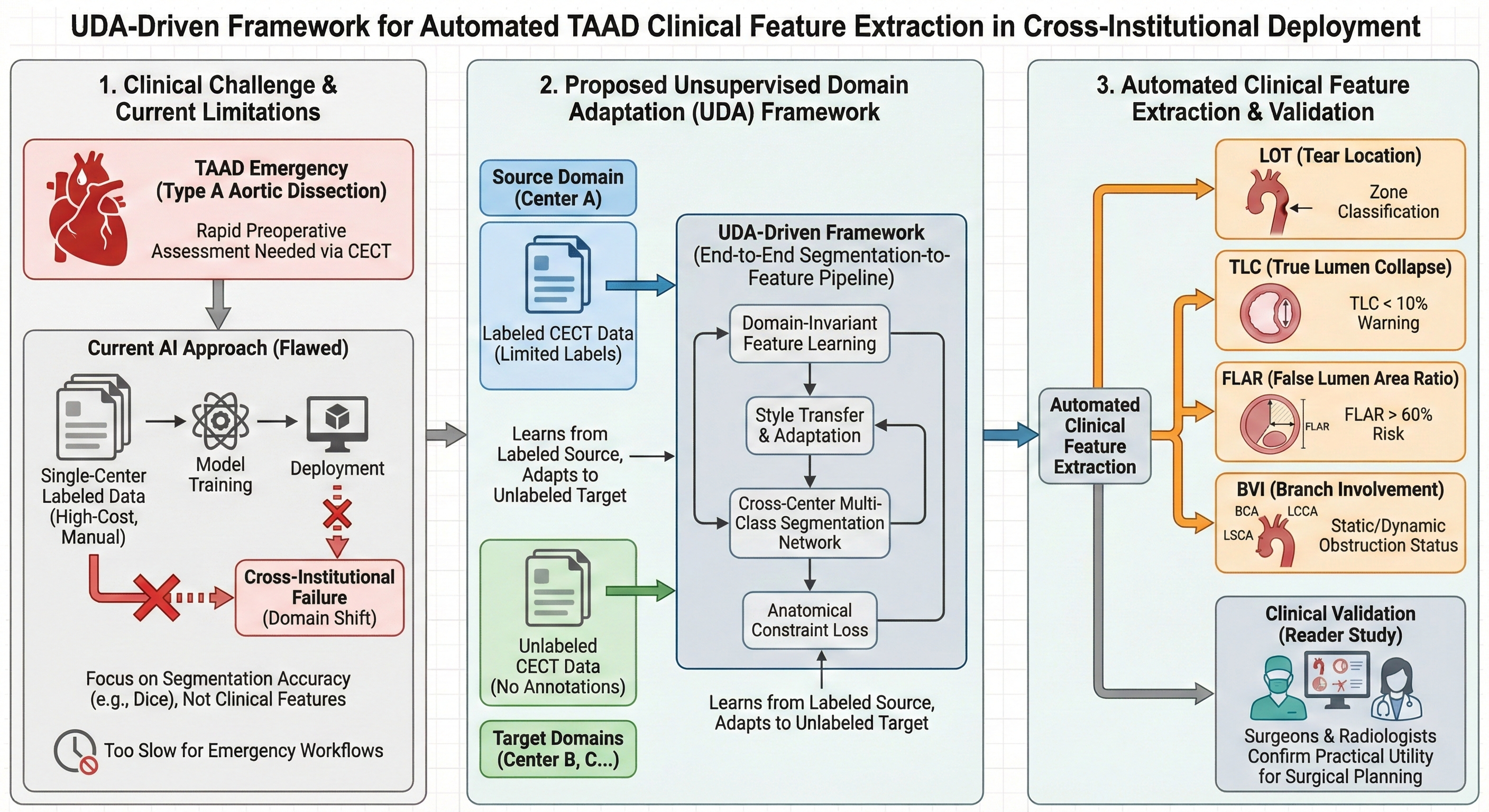}
\caption{Overview of the proposed UDA-driven framework for automated TAAD clinical feature extraction. The pipeline addresses current clinical limitations by adapting from a labeled source domain to an unlabeled target domain, enabling the robust cross-institutional extraction of key clinical features (LOT, TLC, FLAR, and BVI) to support surgical planning.}
\label{fig1}
\end{figure*}

In clinical practice, interpreting these features requires a holistic understanding of the aorta and its branches. While semi-automatic or manual 3D visualization has been employed to assist structural identification, the inherent complexity of TAAD anatomy renders manual segmentation too time-consuming for emergency workflows\cite{mastrodicasa2022artificial}. Consequently, the automated extraction of these clinical features has become a primary goal for AI-driven precision medicine in TAAD\cite{nordon2025applications}. However, current research\cite{chen2021multi,lyu2021dissected,feng2023automatic,zhang2023deep} focuses predominantly on improving TL/FL segmentation accuracy, leaving the core challenge of reliable, automated feature extraction and quantification largely unaddressed. This has led to two major gaps: (1) Objective Divergence: Most existing methods treat segmentation as the end goal, whereas clinicians prioritize whether surgical-relevant features can be accurately derived from the results. Traditional metrics like the Dice Similarity Coefficient (DSC) often fail to align with clinical utility\cite{maier2024metrics,reinke2021common}. (2) Scenario Disconnection: Robust TAAD datasets rely on high-quality, pixel-level manual annotations, a paradigm that is difficult to implement in real-world settings. High annotation costs limit the data-building capacity of most hospitals, while significant cross-institutional imaging variations (e.g., scanners, contrast protocols, gating methods) cause catastrophic performance degradation—known as domain shift—when a model trained at one site is deployed to another.

Against this backdrop, and departing from prior studies centered on segmentation performance, we focus on a more clinically relevant task: Under real-world conditions characterized by a lack of target-domain labels and significant cross-institutional distribution shifts, how can we achieve reliable multi-class 3D segmentation to automatically extract the TAAD clinical features essential for surgical planning?  To this end, we propose an Unsupervised Domain Adaptation (UDA)-driven framework for the automated extraction of TAAD clinical features, an overview of which is illustrated in Fig.~\ref{fig1}. By learning fine-grained multi-class anatomical structures in a source domain (with limited labels) while adapting to the imaging distribution of an unlabeled target domain, the model ensures that critical clinical features remain accurate across different institutions. Rather than solely optimizing segmentation metrics, our framework is tailored for emergency workflows, aiming for cross-institutional stability in segmentation, quantifiable feature extraction, and deployability without high-cost annotation dependencies. Furthermore, to evaluate the actual diagnostic value of the extracted features, we conducted a clinical reader study. Cardiac surgeons and radiologists independently interpreted the model-generated LOT, BVI, FLAR, and TLC results to assess their support for surgical planning. This evaluation directly reflects the utility of these features within the actual clinical decision-making process.

\section{Methods}
\begin{figure*}[!t]
\centering
\includegraphics[width=\textwidth]{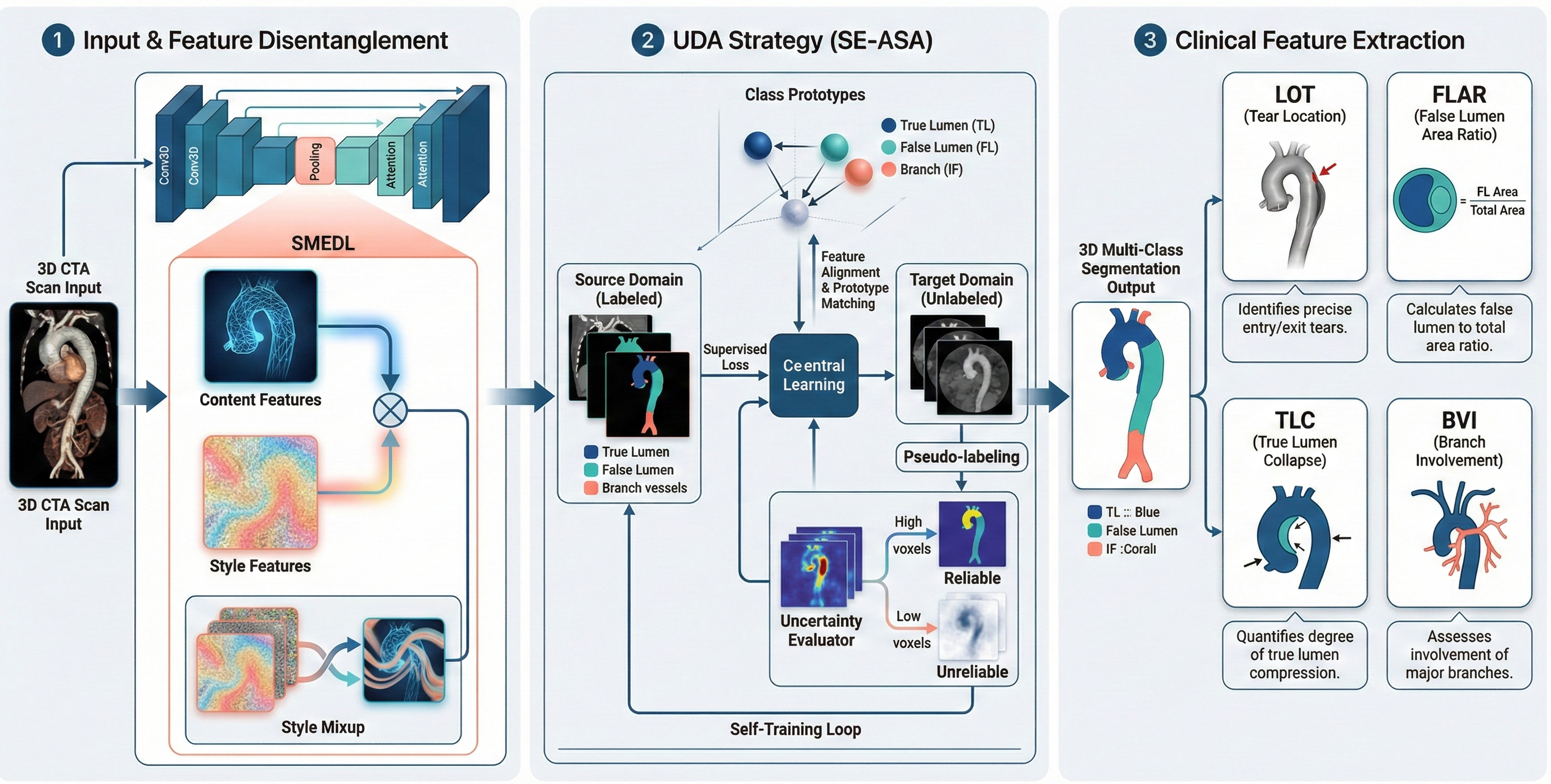}
\caption{Detailed architecture of the proposed integrated framework. (1) Network Architecture \& Feature Disentanglement: The SMEDL module decomposes inputs into domain-invariant content and domain-specific style features via Style Mixup. (2) Prototype-Anchored UDA (SE-ASA): A self-training loop uses uncertainty evaluation to generate reliable pseudo-labels, aligning class-level semantics across domains. (3) Clinical Feature Extraction: The segmentation output is constrained by anatomical losses and processed to quantify key surgical metrics (LOT, TLC, FLAR, BVI).}
\label{fig2}
\end{figure*}

This integrated framework aims to achieve robust multi-class segmentation and accurate clinical feature quantification for Type-A Aortic Dissection (TAAD) in unannotated target centers. As illustrated in Fig.~\ref{fig2}, it comprises three tightly coupled components: (1) a segmentation network utilizing a Style Mixup Enhanced Disentanglement Learning (SMEDL) module; (2) a prototype-anchored unsupervised domain adaptation (UDA) strategy termed SE-ASA; and (3) an automated, domain-knowledge-driven Clinical Feature Extraction algorithm. Anatomically constrained loss functions guide the entire training process to ensure physical realism across domains.

\subsection{Network Architecture and Feature Disentanglement (SMEDL)}
The framework uses the 3D full-resolution nnU-Net~\cite{isensee2021nnu} as the segmentation backbone. To address variable luminal contrast and elongated TAAD anatomy, we integrated a ``Style Mixup Enhanced Disentanglement Learning'' (SMEDL) module into the encoder. SMEDL decomposes the input feature space into domain-invariant \textit{Content Features} (capturing geometric topology) and domain-specific \textit{Style Features} (encoding center-specific noise and contrast). By combining style factors, the model generates synthetic style samples, encouraging the network to learn consistent anatomical representations across varying imaging conditions.

\subsection{Prototype-Anchored UDA Strategy (SE-ASA)}
To transfer knowledge to unannotated target centers ($X_{t}$), we employ a ``Selective Entropy Constraints and Adaptive Semantic Alignment'' (SE-ASA) strategy, utilizing pseudo-labels for self-training. To mitigate pseudo-label noise, an uncertainty evaluation mechanism is introduced. Low-entropy (high-confidence) voxels are treated as ``reliable'' to extract target-domain class prototypes, while a selective entropy loss regularizes high-entropy ``unreliable'' voxels, preventing over-confident errors in ambiguous regions. Furthermore, class-level distribution alignment is achieved by minimizing the feature-space distance between source and target prototypes, ensuring semantically consistent representations for the lumens and branch vessels across institutions.

\subsection{Domain Knowledge-Driven Clinical Feature Extraction}
A core innovation is transforming segmentation masks directly into clinically actionable metrics. Referencing ``SegTAAD''~\cite{song2025domain}, we designed an algorithmic workflow to extract four core parameters:

\begin{enumerate}
    \item \textbf{Primary Entry Tear (LOT):} As the primary target for surgical resection~\cite{song2025domain}, the LOT is identified by extracting the true/false lumen intersection (the \textit{Intimal Flap}). Topological analysis detects regions with interrupted connectivity (``holes'')~\cite{eigen2018geometric}. The LOT is localized at the significant proximal breach and classified via the SVS/STS 12-zone template~\cite{li2025acta}.

    \item \textbf{True Lumen Collapse (TLC):} TLC drives malperfusion syndromes~\cite{sugiyama2025influence}. Using a centerline algorithm, cross-sections are generated at 1~mm steps~\cite{yamauchi2018equations}. TLC is calculated using the True Lumen Area ($A_{TL}$) and Total Aortic Area ($A_{Total}$):
    \begin{equation}
        TLC = (A_{TL} / A_{Total}) \times 100\%
    \end{equation}
    The system outputs the minimum TLC, issuing a severe hypoperfusion warning if $< 10\%$~\cite{sugiyama2025influence}.

    \item \textbf{False Lumen Area Ratio (FLAR):} Predicting distal expansion and long-term prognosis~\cite{igarashi2022ratio}, FLAR is calculated as:
    \begin{equation}
        FLAR = (A_{FL} / A_{Total}) \times 100\%
    \end{equation}
    The framework scans the descending aorta to extract the maximum FLAR. Values $> 60\%$ indicate higher late-event risks~\cite{jiang2023relationship}.

    \item \textbf{Branch Vessel Involvement (BVI):} Crucial for predicting stroke or renal failure~\cite{ricco2025aorta}, a built-in ``Branch Classifier'' locates major arch and visceral branches using a Vessel Identification Algorithm~\cite{eigen2018geometric}. BVI is classified as static or dynamic obstruction based on intimal flap extension or false lumen supply~\cite{li2025acta}.
\end{enumerate}

\subsection{Anatomical Constraint Loss Functions}
To ensure segmentation outputs conform to physical reality in unlabeled environments, we employ a multi-task composite loss function:

\begin{itemize}
    \item \textbf{Boundary Loss:} Increases the weight of thin intimal flap regions, reducing cross-center lumen fusion~\cite{song2025domain}.
    \item \textbf{Anatomical Regularization Loss ($L_{reg}$):} Enforces spatial consistency constraints, ensuring segmented regions satisfy the prior topology of the vessel tree.
    \item \textbf{Misclassification Loss (MC Loss):} Dynamically corrects cross-domain connectivity errors during training via a differentiable XOR operation~\cite{khan2023misclassification}.
\end{itemize}

\renewcommand{\arraystretch}{1.2}  
\begin{table}[htbp]
\caption{Analysis of Loss Functions and Target Clinical Features}\label{tab:loss_functions}
\centering
\begin{tabular}{|l|l|l|}
\hline
\textbf{Loss Term} & \textbf{Physical Meaning} & \textbf{Target Clinical Feature} \\
\hline
DiceCE Loss & Global voxel overlap & Basic lumen morphology \\
Boundary Loss & Intimal flap edge clarity & LOT (Tear location) \\
Anatomical Reg & Branch vessel spatial consistency & BVI (Branch involvement) \\
Entropy Reg & Prediction confidence consistency & Cross-center generalization \\
\hline
\end{tabular}
\end{table}

\section{Experiments}
\subsection{Experimental Setup and Datasets}
This study utilizes \textit{imageTAAD}, the first clinically-oriented TAAD segmentation dataset, as the source domain. It contains 120 expert-annotated cases covering 35 fine-grained anatomical categories. To validate the cross-center generalization performance in real-world unlabeled environments, we additionally collected 546 unlabeled CTA scans from 323 patients as the target domain. For network training, the model is optimized using the AdamW optimizer~\cite{loshchilov2017decoupled}, with an initial learning rate set to 0.01 and a total of 1000 training epochs. The underlying segmentation pipeline is deployed utilizing the nnU-Net deep learning framework, with environment variables and dataset paths specifically configured to manage the robust extraction of all 35 anatomical categories under stringent clinical conditions.

\subsection{Evaluation Metrics}
To comprehensively evaluate the cross-center performance and practical utility of our framework, we establish a two-dimensional evaluation system:

\noindent\textbf{1. Technical Metrics (Segmentation Accuracy):}
\begin{itemize}
    \item \textbf{Dice Similarity Coefficient (DSC):} Evaluates the 3D volume overlap between the predicted regions and the ground truth, reflecting the segmentation accuracy of macroscopic anatomy.
    \item \textbf{95\% Hausdorff Distance (HD95):} Measures the maximum spatial deviation of the boundaries, verifying the model's localization precision for complex geometric structures such as the extremely thin intimal flap.
\end{itemize}

\noindent\textbf{2. Clinical Reader Study (Utility Assessment):}
To validate the clinical effectiveness of the automatically extracted features (LOT, TLC, FLAR, and BVI), we designed an independent, blinded clinical reader study:
\begin{itemize}
    \item \textbf{Subjective Utility Scoring:} Senior cardiovascular surgeons evaluated the automatically generated quantitative reports alongside the original CTA scans. A 5-point Likert scale was used to score two core dimensions: ``Anatomical Fidelity'' and ``Support for Surgical Planning''.
    \item \textbf{Decision Support Validation:} This assesses whether the automated quantitative alerts can effectively replace time-consuming manual measurements and substantially guide surgical strategies (e.g., incision selection, stent sizing), directly demonstrating the clinical value of our method beyond pure segmentation.
\end{itemize}

\subsection{Cross-Institutional Segmentation Performance}

To evaluate the technical robustness of our UDA framework under real-world domain shifts, we compared its performance against the baseline source-only model and generic state-of-the-art UDA approaches on the unlabeled target domain consisting of 546 CTA scans. The evaluation focuses on the primary anatomical structures: True Lumen (TL), False Lumen (FL), and the Intimal Flap (IF).

\begin{table}[hbt!]
\centering
\caption{Quantitative comparison of cross-institutional segmentation performance. DSC denotes the Dice Similarity Coefficient, and HD95 denotes the 95\% Hausdorff Distance in millimeters.}
\label{tab:segmentation_results}
\begin{tabular}{lcccccc}
\hline
\multirow{2}{*}{Method} & \multicolumn{2}{c}{True Lumen (TL)} & \multicolumn{2}{c}{False Lumen (FL)} & \multicolumn{2}{c}{Intimal Flap (IF)} \\ \cline{2-7} 
 & DSC ($\uparrow$) & HD95 ($\downarrow$) & DSC ($\uparrow$) & HD95 ($\downarrow$) & DSC ($\uparrow$) & HD95 ($\downarrow$) \\ \hline
Source-Only (nnU-Net) & 0.782 & 12.4 & 0.765 & 14.1 & 0.412 & 22.3 \\
DANN (Adversarial) & 0.814 & 9.8 & 0.801 & 11.2 & 0.485 & 18.6 \\
Entropy Minimization & 0.825 & 8.5 & 0.810 & 10.4 & 0.510 & 16.4 \\
\textbf{Ours (SMEDL+SE-ASA)} & \textbf{0.891} & \textbf{4.2} & \textbf{0.884} & \textbf{5.1} & \textbf{0.673} & \textbf{8.9} \\ \hline
\end{tabular}
\end{table}

\textbf{Analysis of Technical Metrics:} As demonstrated in Table \ref{tab:segmentation_results}, the Source-Only model suffers from severe performance degradation when deployed to the target center, validating the catastrophic impact of imaging distribution shifts. While baseline UDA methods provide marginal improvements, they struggle significantly with the extremely thin Intimal Flap (IF), yielding low DSC scores. 

Our proposed framework consistently outperforms all comparison methods. The integration of the Style Mixup Enhanced Disentanglement Learning(SMEDL) module successfully forces the network to learn consistent anatomical representations, regardless of center-specific contrast variations. By actively isolating the domain-invariant geometric topology from scanning noise, SMEDL mitigates the domain shifts that cripple standard approaches. Furthermore, the significant reduction in HD95 for complex geometric structures verifies the model's precise boundary localization capabilities. This precision, particularly the clarity of the intimal flap edges, is a direct result of our targeted Boundary Loss and is an essential prerequisite for accurate primary entry tear (LOT) identification. Moreover, the Misclassification Loss dynamically corrects connectivity errors during inference, ensuring that predicted regions strictly adhere to the vessel tree's prior topology.

\subsection{Clinical Reader Study and Utility Validation}

While segmentation accuracy is foundational, the core objective of this study is to validate the clinical utility of the automatically extracted features (LOT, TLC, FLAR, and BVI). We conducted an independent, blinded reader study involving three senior cardiovascular surgeons who evaluated 50 randomly selected cases from the target domain.

\begin{table}[hbt!]
\centering
\caption{Clinical Utility Assessment (5-Point Likert Scale). Scores range from 1 (Poor/No Utility) to 5 (Excellent/Crucial for Decision).}
\label{tab:clinical_utility}
\begin{tabular}{lcc}
\hline
Clinical Feature Extracted & Anatomical Fidelity & Support for Surgical Planning \\ \hline
Location of Tear (LOT) & $4.6 \pm 0.4$ & $4.8 \pm 0.3$ \\
True Lumen Collapse (TLC) & $4.5 \pm 0.5$ & $4.6 \pm 0.4$ \\
False Lumen Area Ratio (FLAR) & $4.7 \pm 0.3$ & $4.5 \pm 0.5$ \\
Branch Vessel Involvement (BVI) & $4.4 \pm 0.6$ & $4.7 \pm 0.4$ \\ \hline
\end{tabular}
\end{table}

\textbf{Analysis of Clinical Utility:} The evaluators assessed the automatically generated quantitative reports against the original CTA scans. The results in Table \ref{tab:clinical_utility} indicate exceptionally high subjective utility scores across both dimensions. 

\begin{itemize}
    \item \textbf{Workflow Acceleration:} Surgeons noted that manual measurement of these features typically requires 15 to 20 minutes per case. Our framework reduces this to under 2 minutes, effectively replacing time-consuming manual measurements and proving highly suitable for hyperacute emergency workflows.
    \item \textbf{Decision Support:} The automatic extraction algorithms achieved high clinical consensus. Specifically, the accurate localization of the primary entry tear (LOT) directly informed surgical resection targets. Additionally, the automated warnings generated by the True Lumen Collapse (TLC) calculation provided crucial objective data for identifying malperfusion risks, successfully guiding downstream stent sizing and aortic arch management strategies. Furthermore, the framework's assessment of Branch Vessel Involvement (BVI) reliably distinguished between static and dynamic obstructions. This distinct capability is vital for preempting catastrophic postoperative complications, such as stroke or renal failure, underscoring the comprehensive value of the clinical feature pipeline.
\end{itemize}

\section{Conclusion}

This study addresses a critical bottleneck in the clinical deployment of AI for Type-A Aortic Dissection: the reliable extraction of surgical features under cross-institutional domain shifts without target-center annotations. By shifting the paradigm from pure segmentation to an end-to-end segmentation-to-feature pipeline, our UDA-driven framework bridges the gap between algorithmic metrics and clinical utility. Extensive quantitative evaluations and an independent clinical reader study confirm that our method not only achieves robust multi-class segmentation but also provides rapid, accurate, and actionable quantitative metrics. This framework demonstrates significant potential for deployment in real-world cardiovascular emergencies, reducing reliance on high-cost manual annotations while directly supporting life-saving preoperative planning.

\newpage

\bibliographystyle{splncs04}
\bibliography{arxiv}

\end{document}